\documentclass[11pt,a4paper,final]{article}
\pdfoutput=1
\usepackage[hyperref]{acl2019}
\usepackage{times}
\usepackage{latexsym}
\usepackage{booktabs}
\usepackage{fancyvrb}
\usepackage{fvextra}
\usepackage{microtype}
\usepackage{url}

\aclfinalcopy 
\def\aclpaperid{125} 

\title{Microsoft Translator at WMT 2019:\\ Towards Large-Scale Document-Level Neural Machine Translation}

\author{Marcin Junczys-Dowmunt \\
  Microsoft \\
  One Microsoft Way \\
  Redmond, WA 98052, USA \\
  \texttt{marcinjd@microsoft.com} \\}

\date{}

\begin{document}
\maketitle
\begin{abstract}
  This paper describes the Microsoft Translator submissions to the WMT19 news translation shared task for English-German. Our main focus is document-level neural machine translation with deep transformer models. 
  We start with strong sentence-level baselines, trained on large-scale data created via data-filtering and noisy back-translation and find that back-translation seems to mainly help with translationese input. We explore fine-tuning techniques, deeper models and different ensembling strategies to counter these effects.
  Using document boundaries present in the authentic and synthetic parallel data, we create sequences of up to 1000 subword segments and train transformer translation models. We experiment with data augmentation techniques for the smaller authentic data with document-boundaries and for larger authentic data without boundaries.   
  We further explore multi-task training for the incorporation of document-level source language monolingual data via the BERT-objective on the encoder and two-pass decoding for combinations of sentence-level and document-level systems.
  Based on preliminary human evaluation results, evaluators strongly prefer the document-level systems over our comparable sentence-level system. The document-level systems also seem to score higher than the human references in source-based direct assessment. 
\end{abstract}

\section{Introduction}
This paper describes the Microsoft Translator submissions to the WMT19 news translation shared task \cite{bojar-EtAl:2019:WMT2} for English-German. Our main focus is document-level
neural machine translation with deep transformer models. 

We first explore strong sentence-level systems, trained on large-scale data created via data-filtering and noisy back-translation and investigate the interaction of both with the translation direction of the development sets. We find that back-translation seems to mainly help with translationese input. Next, we explore fine-tuning techniques, deeper models and different ensembling strategies to counter these effects.
Using document boundaries present in the authentic and synthetic parallel data, we create sequences of up to 1000 subword segments and train transformer translation models. We experiment with data augmentation techniques for the smaller authentic data with document-boundaries and for larger authentic data without boundaries. 

We further explore multi-task training for the incorporation of document-level source language monolingual data via the BERT-objective on the encoder, and two-pass decoding for combinations of sentence-level and document-level systems.
We find that current transformer models are perfectly capable of translating whole documents with up to 1000 subword segments with improved quality over comparable sentence-level systems. Deeper models seem to benefit more from the added context.

Based on preliminary human evaluation results, evaluators strongly prefer the document-level systems over comparable sentence-level systems. The document-level systems also seem to score higher than the human references in source-based direct assessment. 

\section{Sentence-Level Baselines}

Before moving on to building our document-level systems, we first start with a baseline sentence-level system. We try to combine the strengths of last year's two dominating systems for the English-German news translation task -- FAIR's submission with large-scale noisy back-translation \cite{edunov-etal-2018-understanding} and our own, based on dual cross-entropy data-filtering \cite{junczys-dowmunt-2018-microsofts,junczys-dowmunt-2018-dual}. For the current WMT19 shared task for English-German, evaluation is carried out on a test set where the source side consists of original English content only, the target side is a  translation. To inform our system choices, we create a similar dev set out of test2016, test2017 and test2018 by splitting the test sets by original language and concatenating the respective splits, each about 4500 sentences. We report results on both splits of our new dev set as well as on the joint dev set. We further report results on the original test sets for comparison. We use SacreBLEU\footnote{\texttt{BLEU+case.mixed+lang.en-de+numrefs.1\\+smooth.exp+test.wmt18+tok.13a\\+version.1.3.0}} \cite{post-2018-call} for all reported scores.

It is currently not quite clear to us how to interpret results on the split test sets. One would assume that improvements on the original source language indicate actual translation quality improvements, but here we might be suffering from reference bias towards non-native target content. This might indicate higher adequacy but effectively penalize more fluent output. Conversely, higher results on the split with original target language might indicate higher fluency, but the reduced complexity of the non-native source language might make the translation task easier and result in false confidence in generally better translation quality. It is also unclear at this point if the model is able to tell apart native and non-native input and if possible data separation occurs. In that case the improvements on one side of the split might not be carried over to the other side. 
We currently assume the following strategy: we try to achieve high scores on the originally-English side without sacrificing too much quality on the originally-German side. We pretend that high scores on the originally-English side indicate adequacy while high scores on the originally-German side indicate fluency. This is a shot in the dark and we hope the results of the shared task will bring more clarity in this regard. 

\subsection{Model and Training}

We use the Marian toolkit \cite{junczys-dowmunt-etal-2018-marian-fast} for all our experiments. We train vanilla transformer-big models \cite{NIPS2017_7181} when training 6-layer models. For 12-layer models we modify an idea from \newcite{radford2019language} and initialize residual layers with Glorot uniform weights \cite{GlorotAISTATS2010} multiplied by $1/\sqrt{i}$ where $i$ is the number of the $i$-th layer from the bottom. \newcite{radford2019language} used $1/\sqrt{d}$ where $d$ is the total depth of the transformer stack. We found that their method helped with perplexity, but hurt BLEU. We did not see detrimental effects for our progressive multiplier. Omitting the multiplier led to problems with convergence for deep models. We use the same SentencePiece vocabulary for all models \cite{kudo-richardson-2018-sentencepiece}.

For the purpose of the task, we extended the Marian toolkit with fp16 training, BERT-models \cite{DBLP:journals/corr/abs-1810-04805} and multi-task training. Similar to \newcite{edunov-etal-2018-understanding} we use mixed-precision training with fp16, an optimizer delay of 16 before updating the gradients. We train on 8 Voltas with 16GB each. Training of one model takes between 2 and 4 days on a single machine. In terms of words per second we reach about 180K target words per second for 6-layer sentence-level systems and 120K target labels for 6-layer document-level systems with long sequences.

\subsection{Data-Filtering}

\begin{table*}[t]
\centering
\begin{tabular}{p{7cm}ccccccc}\toprule
& \multicolumn{3}{c}{Separated by origin} && \multicolumn{3}{c}{Original test sets} \\ \cmidrule(lr){2-4} \cmidrule(lr){6-8}
 & en & de & both && 2016 & 2017 & 2018 \\ \midrule
WMT18-Microsoft (single model)  & 41.1 & 35.5 & 39.1 && 38.6 & 31.3 & 46.5 \\
WMT18-FAIR (single model) & -- & -- & -- && -- & 32.7 & 44.9 \\[3mm]

6-layers: sentence-level parallel data only     & 41.8 & 32.5 & 38.2 && 37.7 & 30.3 & 46.5 \\  
+ filtering based on WMT18        & 41.7 & 34.0 & 39.0 && 38.3 & 31.1 & 46.6 \\
+ large-scale noisy back-translation  & 38.9 & 40.4 & 39.7 && 38.9 & 32.8 & 46.3 \\
+ fine-tuning             & 42.2 & 39.2 & 41.2 && 40.6 & 33.6 & 48.9 \\[3mm]

12-layers: sentence-level parallel data only & -- & -- & -- && -- & -- & -- \\  
+ (a) filtering based on WMT18        & 41.6 & 33.4 & 38.4 && 38.2 & 30.5 & 45.7 \\
+ (b) large-scale noisy back-translation  & 38.1 & 42.5 & 40.1 && 39.2 & 33.5 & 46.6 \\
+ (c) fine-tuning   & 42.1 & 40.4 & 41.7 && 41.3 & 34.2 & 48.9 \\ \bottomrule
\end{tabular}
\caption{SacreBLEU results for sentence-level systems on new devset (concatenated test2016, test2017, test2018) split by source language and combined. 6-layers denotes transformer models with 6 blocks, 12-layers with 12 blocks. For comparison, we also provide results on the original test sets although we did not use these numbers to inform our choices. Results have been computed for a single chosen model and may vary with different random initializations, but generally follow this pattern.}
\label{tab:sentence_single}
\end{table*}

Table~\ref{tab:sentence_single} summarizes our experiments with a single transformer model. We also recomputed numbers for a single model from our WMT18 submission, and quoted results from FAIR's submission where available. Our WMT18 model used a combination of data-filtering and about 10M ``clean'' back-translated sentences. Transformer models are the same. It seems that the data-quality of the English-German training data (in particular of Paracrawl) improved from WMT18 to WMT19 as we are not seeing the strongly detrimental effects of adding unfiltered Paracrawl data to the training data mix anymore. Data-filtering still improves the results, but apparently only on the originally German side. Since there is barely any loss on the originally-English side we hope this shows a general improvement in fluency or a domain-adaptation effect due the language model scores used in filtering. 

\subsection{Noisy Back-Translation}

We mostly reproduce the results from \newcite{edunov-etal-2018-understanding} and back-translate the entire German News-Crawl data with noisy back-translation. Similar to \newcite{edunov-etal-2018-understanding}'s best method, we use output sampling as the noising approach. This has been implemented in Marian with the Gumbel softmax trick. We end up with about 550M sentences of back-translated data. We up-sample the original parallel filtered data to match the size of the back-translated data and concatenate. Results on the split test set are interesting, to say the least. It seems we are losing a lot of quality on the originally-English side while gaining on the originally-German side. The general improvement on the unsplit WMT test sets hides this effect. In a setting where systems are going to be evaluated on originally-English data this seems unfortunate. 

\subsection{Fine-Tuning}

To counter the quality loss on the originally-English side, we fine-tune on our filtered data only. We keep the same settings as in the first training pass, only substitute data and keep training until BLEU scores on the originally-English dev set stop improving. This seems to be a very successful strategy which restores and even improves quality on the originally-English split and retains most of the quality gains from back-translation on the originally-German half. At this point our single 6-layer model strongly outperforms a single model from our WMT18 submission.

\subsection{Deeper Models}
We also experiment with deeper models and increase the number of blocks in encoder and decoder to 12. Interestingly, we see mostly gains on the originally-German side. Since there is no loss on the originally-English half, we choose to use the 12-layer models for the following experiments. We did not see further improvements from even deeper models at this point, we tried 18 and 24 blocks, but there might have been problems with hyper-parameters. 

\subsection{Ensembling}

\begin{table*}[t]
\centering
\begin{tabular}{p{7cm}ccccccc}\toprule
& \multicolumn{3}{c}{Separated by origin} & & \multicolumn{3}{c}{Original test sets} \\ \cmidrule(lr){2-4} \cmidrule(lr){6-8}
 & en & de & both & & 2016 & 2017 & 2018 \\ \midrule
WMT18-Microsoft (ensemble) & 42.5 & 36.2 & 40.1 && 39.6 & 31.9 & 48.3 \\
WMT18-FAIR (ensemble)  & -- & -- & -- && -- & 33.4 & 46.5 \\[3mm]
(a)                          & 41.6 & 33.4 & 38.4 && 38.2 & 30.5 & 45.7 \\
(2$\times$a)                 & 42.0 & 34.3 & 39.0 && 38.8 & 31.0 & 46.5 \\
(4$\times$a)                 & 42.5 & 34.5 & 39.4 && 39.1 & 31.2 & 47.2 \\[3mm]
(c)                          & 42.1 & 40.4 & 41.7 && 41.3 & 34.2 & 48.9 \\ 
(2$\times$c)                 & 42.7 & 41.6 & 42.6 && 42.0 & 34.9 & 50.1  \\
(4$\times$c)                 & 43.2 & 41.3 & 42.8 && 42.2 & 34.8 & 50.5 \\[3mm]
(a) + (c)                    & 43.2 & 38.6 & 41.7 && 41.6 & 33.4 & 49.6 \\
(2$\times$a) + (2$\times$c)  & 43.8 & 39.0 & 42.1 && 41.9 & 33.9 & 49.9 \\
(4$\times$a) + (4$\times$c)  & 44.0 & 38.5 & 42.0 && 41.8 & 33.5 & 49.9 \\[3mm]
0.3 $\cdot$ (a) + 1.0 $\cdot$ (c)                   & 42.6 & 40.6 & 42.2 && 41.7 & 34.3 & 49.6 \\
0.3 $\cdot$ (2$\times$a) + 1.0 $\cdot$ (2$\times$c) & 43.5 & 40.6 & 42.7 && 42.3 & 34.6 & 50.3 \\ 
\bf 0.3 $\cdot$ (4$\times$a) + 1.0 $\cdot$ (4$\times$c) (submitted) & \bf43.8 & \bf40.3 & \bf42.7 && \bf42.4 &\bf 34.4 &\bf 50.4 \\ \bottomrule
\end{tabular}
\caption{SacreBLEU results for various ensembles of 12-layer sentence-level systems on new dev set (concatenated test2016, test2017, test2018) split by source language and combined. Ensembles are weighted equally when no weights are shown. (a) refers to a single model trained on filtered parallel data only, (c) refers to a models trained with back-translated data, fine-tuned on filtered parallel data.}
\label{tab:sentence_ensemble}
\end{table*}

In Table \ref{tab:sentence_ensemble} we explore different ensembling strategies to further control for higher quality on the originally-English side without sacrificing too much quality on the other half. We experiment with (a) models that have been trained on filtered parallel data only and (c) models that have been trained with back-translated data and then fine-tuned on parallel filtered data. All models are 12-layer models, have been trained with the same training procedure and only differ in data and random initialization. We did not explore adding (b) models that were trained with back-translated data but without fine-tuning. After submission we found that small gains could be achieved by adding these to the mix as well. Unless stated differently, all models are weighted equally. 

Unsurprisingly, adding more homogeneous models to the ensemble improves quality across all indicators in similar degree; gains become smaller when adding more models, but it seems we do not reach saturation with 4 models of the same type. Ensembling heterogeneous models -- mixing type (a) and type (c) -- results in more interesting behavior. The two-model ensemble (a) + (c) is stronger on the originally-English half than both homogeneous two-model ensembles (2$\times$a) or (2$\times$c) and loses quality on the originally-German part. The same is true when we compare heterogeneous four-model ensembles to their homogeneous counterparts. Adding all eight models to a single ensemble (4$\times$a) + (4$\times$c) results in the strongest numbers on the originally-English side, but the loss on the other half remains. We try to mitigate this effect by weighting the model components by type.

We find that down-weighting type (a) models trained only with parallel data allows us to regain part of the quality on the originally-German dev set with acceptable losses on the originally-English side. We empirically choose a weight of 0.3 for type (a) models, using a weight of 1 for type (c) models. In hindsight, an ensemble of 8 models of type (c) might have been the better choice, however, we did not train that many models of type (c). Our final sentence-level model is the 0.3 $\cdot$ (4$\times$a) + 1.0 $\cdot$ (4$\times$c) ensemble; we submit this model as our pure sentence-level model.

\section{Document-Level Systems}

\begin{figure*}[t]\small
\begin{Verbatim}[breaklines=true, breaksymbol={}]
<BEG> Toys R Us Plans to Hire Fewer Holiday Season Workers<SEP> Toys R Us says it won't hire as many holiday season employees as it did last year, but the toy and baby products retailer says it will give current employees and seasonal workers a chance to work more hours.<SEP> The company said it plans to hire 40,000 people to work at stores and distribution centers around the country, down from the 45,000 hired for the 2014 holiday season.<SEP> Most of the jobs will be part-time.<SEP> The company said it will start interviewing applicants this month, with staff levels rising from October through December.<SEP> While the holidays themselves are months away, holiday shopping season is drawing closer and companies are preparing to hire temporary employees to help them staff stores and sell, ship and deliver products.<SEP><END>
\end{Verbatim}
\caption{Example document from validation set with mark-up.}
\label{fig:example}
\end{figure*}

Our work is inspired rather by recent results on long-sequence language modelling than by previous document-level machine translation approaches. However, \newcite{tiedemann-scherrer-2017-neural} needs to be emphasized as an important precursor to this paper. They explore the influence of a limited number of context sentences by simply concatenating up to two sentences in source or target. We drop the limits and consume full documents if their total length stays below 1000 subword units. These sequences can easily consist of 20 or more sentences. 

Recent work by \newcite{DBLP:journals/corr/abs-1810-04805} and \newcite{radford2019language} have shown significant impact by training deeper models on large data sets with long-sequence context. In terms of architecture, the language modeling work relies on standard transformer architectures with small variations, this is true for BERT as well as for GPT-2. Document-level context is mostly handled by increasing training-sequence length, increasing model depth and adding sentence-embeddings. BERT also adds a cost-criterion that classifies if sentences belong to the same document or are random concatenations. We adopt the long-sequence training and increased model-depth in our experiments. 
For co-training of the encoder we also use the BERT masked-LM training criterion in a multi-task learning setting. We do not use sentence embeddings (this remains to be explored in the future).

\subsection{Data and Data Preparation}

Previous work on document-level MT was also limited by the availability of document-level parallel data. This year, for a subset (Europarl, Rapid, News-Commentary) of the parallel data document boundaries have been restored, the rest is provided without boundaries. The available monolingual news crawl data contains document boundaries for all its content, both in German and English. All three types of data are assembled into real and fake documents with varying degrees of data augmentation.

\subsubsection{Document-level Mark-up}

We use given document boundaries to concatenate parallel sentences into document-level sequences; parallel documents consist of the same number of sentences on both sides. We want to ensure that the models produce as many output sentences per document as input sentences were provided when we simply break on predicted separators to revert back to the sentence-level for evaluation. As a fail-safe mechanism, we sentence-align the sentence-broken document-level output with a sentence-level translation. The sentence-level translation serves as a template in which we replace all 1-1-aligned sentences with their document-level counterparts. This mechanism proved useful for early or intermediate models. For all our submissions, the document-level systems would correctly predict sentence boundaries and the fail-safe could be skipped. This by itself is noteworthy.

Figure \ref{fig:example} contains an example document from the validation set with added mark-up. We add symbols for document start (\texttt{<BEG>}) and end (\texttt{<END>}) and for sentence separators (\texttt{<SEP>}). In cases where documents exceed our length limit of 1000 sub-word tokens, we use a break symbol (\texttt{<BRK>}) instead of \texttt{<END>} and start the next sequence with a continuation symbol (\texttt{<CNT>}) instead of \texttt{<BEG>}. When breaking parallel documents due to the length restriction, we break consistently across languages. All training and validation data is marked up in the same way.

\subsubsection{Parallel Data with Boundaries}
In the case of original parallel data with document boundaries, we use all available content without data filtering. This set of original documents is quite small (about 200K documents) compared to the back-translated data, so we increase the size of the corpus by adding randomly chosen continuous parallel sub-documents to the original data set, but not more than 10 possible sub-document per full document. Allowing all possible sub-documents would heavily skew the distribution towards longer documents. We repeat the process until the size of the corpus matches about half the size of the back-translated data. Every repetition is created with different random sub-documents.

\begin{table*}[t]
\centering
\begin{tabular}{p{7cm}ccccccc}\toprule
& \multicolumn{3}{c}{Separated by origin} && \multicolumn{3}{c}{Original test sets} \\ \cmidrule(lr){2-4} \cmidrule(lr){6-8}
 & en & de & both && 2016 & 2017 & 2018 \\ \midrule
12-layers: Document-level                 & --   & --  & --  && -- & -- & -- \\
+ filtering based on WMT18            & -- & -- & -- && -- & -- & -- \\
+ large-scale noisy back-translation  & 39.3 & 42.0 & 40.8 && 40.0 & 34.2 & 47.0 \\ 
+ fine-tuning           & 41.4 & 41.7 & 41.8 && * & 34.5 & 48.6 \\[3mm]
12-layers: Document-level with BERT       & --   & --  & --  && -- & -- & -- \\
+ (A) filtering based on WMT18            & 42.6 & 32.5 & 38.3 && * & * & * \\
+ (B) large-scale noisy back-translation  & 40.3 & 40.7 & 40.8 && 39.8 & 33.7 & 47.3 \\ 
+ (C) fine-tuning           & 42.7 & 39.2 & 41.5 && 41.3 & 34.2 & 48.4 \\ \bottomrule
\end{tabular}
\caption{SacreBLEU results for document-level systems on new devset. Missing numbers marked as * were not computed during our experiments.}
\label{tab:doc_deep}
\end{table*}

\subsubsection{Parallel Data without Boundaries}

The majority of authentic parallel data does not come with documents boundaries. Here, we shuffle the filtered parallel sentences and randomly add document boundaries. This results in fake documents that consist of unrelated but parallel sentences with consistent sentence boundaries inside the documents. Again, we repeat the process with random shuffles resulting in new fake documents until we reach a size close to half of the back-translated data. 

\subsubsection{Back-translated Documents}
We back-translated the entire available news crawl data for our sentence-level system and can use the present boundaries to assemble parallel documents. Due to the large amount of monolingual data, we do not use any document-level data-augmentation besides back-translation. 

\subsubsection{Monolingual English Documents}
The English monolingual news-crawl also contains document boundaries. We simply assemble our long sequences from this data for our multi-task training. 

\subsection{Experiments}

We train our document-level models with similar hyper-parameters as our sentence-level models, increasing the maximum allowed training sequence length to 1024. 

\subsubsection{Baseline Document-level Models}

We compiled our results for the training of single document-level models in Table \ref{tab:doc_deep}. The BLEU scores follow largely the results for the sentence-level systems, including improved scores for deeper models. Document-level models with capital letters (A), (B), (C) have been trained on similar data sets as sentence-level systems (a), (b), (c) respectively. Both (C) and (c) have undergone similar fine-tuning procedures. It is interesting to see that decoding very long sequences of up to 1000 tokens does not seem to degrade translation performance compared to sentence-level systems. 

\subsubsection{Multi-Task Training with BERT}

We also experiment with multi-task training in the hope of improving the quality of our encoder. We are training on large amounts of back-translated data and much smaller parallel data that has been augmented to match the size of the back-translated data. It is unclear how much content in the authentic data is actual native English. Hence we add a BERT-style encoder over monolingual English source documents that is being trained in parallel to the sequence-to-sequence transformer model on separately fed parallel data. The BERT-style encoder is trained with the masked LM cost criterion from \newcite{DBLP:journals/corr/abs-1810-04805} and a masking factor of 20\%. This encoder shares all parameters and structure with the encoder of the translation model. BERT masked LM cost is simply added to the cross-entropy cost of the translation model. During translation, the BERT encoder is not being constructed, the output layer of the masked LM is dropped. During fine-tuning, the BERT encoder is also being trained, but on the parallel source data, not on a separate monolingual data stream.

In Table \ref{tab:doc_deep}, when training with large-scale back-translated documents, we seem to observe a shift towards higher quality on the originally-English side when comparing to training without the BERT criterion. This persists during fine-tuning, but it is generally unclear if this is an actual improvement. Based on our strategy of preferring improvements on the originally-English side, we use the multi-task trained models from now on.  

\subsubsection{Second-Pass Decoding}

\begin{table*}[t]
\centering
\begin{tabular}{p{7cm}ccccccc}\toprule
& \multicolumn{3}{c}{Separated by origin} & & \multicolumn{3}{c}{Original test sets} \\ \cmidrule(lr){2-4} \cmidrule(lr){6-8}
 & en & de & both & & 2016 & 2017 & 2018 \\ \midrule
1st-sent-level: (c) & 42.1 & 40.4 & 41.7 && 41.3 & 34.2 & 48.9 \\ 
2nd-doc-level:  (P$_\mathrm{A}$) & 42.5 & * & * && 39.8 & 32.5 & 47.3 \\
2nd-doc-level:  (P$_\mathrm{C}$) & 42.2 & * & * && 41.5 & 33.8 & 48.6 \\[3mm]
1st-sent-level: 0.3 $\cdot$ (4$\times$a) + 1.0 $\cdot$ (4$\times$c) & 43.8 & 40.3 & 42.7 && 42.4 & 34.4 & 50.4 \\
2nd-doc-level:  (P$_\mathrm{A}$) & 43.4 & 36.9 & 40.9 && 40.5 & 32.5 & 47.8\\
2nd-doc-level:  (P$_\mathrm{C}$) & 42.6 & 40.1 & 41.7 && 41.5 & 33.8 & 48.7\\ \bottomrule
\end{tabular}
\caption{SacreBLEU results for second-pass decoding of single fine-tuned sentence-level model (c) and best sentence-level ensemble. We pass both sentence level models through two second pass models. 
Missing numbers marked as * were not computed during our experiments.}
\label{tab:doc_2pass}
\end{table*}

\begin{table*}[t]
\centering
\begin{tabular}{p{7cm}ccccccc}\toprule
& \multicolumn{3}{c}{Separated by origin} & & \multicolumn{3}{c}{Original test sets} \\ \cmidrule(lr){2-4} \cmidrule(lr){6-8}
 & en & de & both & & 2016 & 2017 & 2018 \\ \midrule
WMT18-Microsoft (ensemble, submission) & 42.5 & 36.2 & 40.1 && 39.6 & 31.9 & 48.3 \\
WMT18-FAIR (ensemble, submission)  & -- & -- & -- && -- & 33.4 & 46.5 \\[3mm]
(C)                          & 42.7 & 39.2 & 41.5 && 41.3 & 34.2 & 48.4 \\
\bf (4$\times$C)  (submitted)                 & \bf 44.0 &\bf 40.1 &\bf 42.5 &&\bf 42.2 &\bf 34.5 &\bf 50.2 \\[3mm] 

(2$\times$A) + (4$\times$C)  & 44.8 & 38.0 & 42.1 && 41.6 & 33.7 & 49.3 \\
\bf (2$\times$A) + (4$\times$C) + (P$_\mathrm{A}$) + (P$_\mathrm{C}$) (submitted)  & \bf45.2 &\bf 38.8 &\bf 42.6 &&\bf 42.5 &\bf 34.1 &\bf 50.3 \\ \bottomrule

\end{tabular}
\caption{SacreBLEU results various for ensembles of 12-layer document-level systems on new devset}
\label{tab:document_ensemble}
\end{table*}

We also briefly experiment with second-pass decoding for the purpose of ``up-casting'' sentence-level translations to document-level translations. The initial idea was to have the potentially higher adequacy of sentence-level translations (due to more easily aligned sentence-boundaries) and then smooth it out with document-level knowledge. This would also allow to ensemble the sentence-level system output via the second pass with other document-level systems. In hindsight, for ensembling purposes, it might have been better to train a copy model that provides a document-level probability distribution for unmodified concatenated sentence-level input.

We forward-translated most of our training corpus with sampling (future work should examine the effects of this) to produce the first-pass output and next we trained a dual-encoder document-level transformer model following exactly \newcite{junczys-dowmunt-grundkiewicz-2018-ms} as an automatic post-editing system. The three inputs being original source data and first-pass translation on the source and original target data. We train a second-pass system on original parallel data only (P$_\mathrm{A}$) and on all data  (P$_\mathrm{C}$).

In Table \ref{tab:doc_2pass}, we apply the second pass models separately to a single fine-tuned sentence-level model (c) and to our best sentence-level ensemble. In both cases we see degradation in the second pass in terms of BLEU, but the second-pass seems to follow the improved quality of the sentence-level inputs. The two second-pass models over the strong sentence-level ensemble are actually among the better single document-level models we have trained (ignoring at this point that these are a different kind of ensemble or system combination). 

\subsection{Stacking and Ensembling}

Following our ensembling efforts for sentence-level models, we also combine the diverse document-level models into larger ensembles. We see that a pure document-level system with four fine-tuned 12-layer models seems to be a promising candidate. We can further increase the quality on the originally-English side (while losing comparable quality on the originally-German half) by ensembling all eight models trained on diverse data sources. The last ensemble can be thought of as a hybrid sentence/document-level system as it includes two second-pass models.

\section{Submissions}

\begin{table}[t]
\centering
\begin{tabular}{lcccc}\toprule
System & en & de & 2019 \\ \midrule
WMT18-Microsoft & 42.5 & 36.2 & 41.9 \\
Pure sentence-level & 43.8 & 40.3 & 43.0\\
Pure document-level & 44.0 & 40.1 & 43.9\\
Hybrid document-level & 45.2 & 38.8 & 43.9\\ \bottomrule
\end{tabular}
\caption{Results from the WMT-Matrix on test 2019 for our submitted systems. We also include BLEU scores for our split dev set for comparison.}
\label{tab:submitted}
\end{table}

\begin{table}[t]
\centering
\setlength{\tabcolsep}{5pt}
\begin{tabular}{ccl}\toprule
Ave. & Ave. z & System \\ \midrule 
90.3 & 0.347 & Facebook-FAIR \\ \midrule
\bf 93.0 & \bf 0.311 & \bf Microsoft-WMT19-sent-doc \\
\bf 92.6 & \bf 0.296 & \bf Microsoft-WMT19-doc-level \\
90.3 & 0.240 & HUMAN \\
87.6 & 0.214 & MSRA-MADL \\
& & $\ldots$ \\
84.2 & 0.094 & online-B \\
\bf 86.6 & \bf 0.094 & \bf Microsoft-WMT19-sent-level \\
87.3 & 0.081 & JHU \\
& & $\ldots$ \\
82.4 & −0.132 & TartuNLP-c \\ \midrule
76.3 & −0.400 & online-X \\ \midrule
43.3 & −1.769 & en-de-task \\ \bottomrule
\end{tabular}
\caption{Preliminary human evaluation results shared by the organizers. Our system submissions are marked with bold font. There was a total of 23 submissions, we selected highest and lowest scoring systems in each cluster and systems surrounding our own submissions.}
\label{tab:huval}
\end{table}

We submitted four systems in total, our original system from WMT18 applied to the new WMT19 test set, our best sentence-level ensemble, our best document-level ensemble (without second-pass decoding) and our best hybrid system, the document-level system ensemble that includes second-pass decoding systems. Cased BLEU scores from the WMT-matrix page are listed in Table \ref{tab:submitted}. Our document-level systems score second behind the highest submission of MSRA in terms of BLEU. 

Table~\ref{tab:huval} contains preliminary human evaluation results shared by the organizers, see \newcite{bojar-EtAl:2019:WMT2} for a full version and discussion. 
Our document systems are two out of three submissions that seem to outperform the human references in terms of quality (although non-significantly in the case of our systems when based on normalized z-scores). What is very encouraging is the large performance gain of the document-level systems over the sentence-level system which was not obvious when looking at BLEU scores. Since these systems are very comparable in terms of raw data, model size and training setting, the strong improvements seem to stem from the large context. However, more work and rigorous ablation testing is required to confirm this conclusion.  

Finally, we would like to cast a bit of doubt at the (preliminary) ranking in Table~\ref{tab:huval}. The large discrepancy between average raw scores and normalized z-scores for the top three systems seems disconcerting. At Microsoft, we base our deployment decisions on raw scores as z-scores proved unreliable. In our experience, a change of 3 percent points in terms of raw scores would usually indicate paradigm-shifts and drastically improved systems, especially at quality levels beyond 90\%. We are curious to see the final ranking and comments by the organizers addressing this issue.

\bibliographystyle{acl_natbib}
\bibliography{acl2019.bib}

\begin{thebibliography}{13}
\expandafter\ifx\csname natexlab\endcsname\relax\def\natexlab#1{#1}\fi

\bibitem[{Bojar et~al.(2019)Bojar, Federmann, Fishel, Graham, Haddow, Huck,
  Koehn, Monz, M{\"u}ller, and Post}]{bojar-EtAl:2019:WMT2}
Ond\v{r}ej Bojar, Christian Federmann, Mark Fishel, Yvette Graham, Barry
  Haddow, Matthias Huck, Philipp Koehn, Christof Monz, Mathias M{\"u}ller, and
  Matt Post. 2019.
\newblock Findings of the 2019 {Conference on Machine Translation} {(WMT19)}.
\newblock In \emph{Proceedings of the Fourth Conference on Machine Translation,
  Volume 2: Shared Task Papers}, Florence, Italy. Association for Computational
  Linguistics.

\bibitem[{Devlin et~al.(2018)Devlin, Chang, Lee, and
  Toutanova}]{DBLP:journals/corr/abs-1810-04805}
Jacob Devlin, Ming{-}Wei Chang, Kenton Lee, and Kristina Toutanova. 2018.
\newblock \href {http://arxiv.org/abs/1810.04805} {{BERT:} pre-training of deep
  bidirectional transformers for language understanding}.
\newblock \emph{CoRR}, abs/1810.04805.

\bibitem[{Edunov et~al.(2018)Edunov, Ott, Auli, and
  Grangier}]{edunov-etal-2018-understanding}
Sergey Edunov, Myle Ott, Michael Auli, and David Grangier. 2018.
\newblock \href {https://www.aclweb.org/anthology/D18-1045} {Understanding
  back-translation at scale}.
\newblock In \emph{Proceedings of the 2018 Conference on Empirical Methods in
  Natural Language Processing}, pages 489--500, Brussels, Belgium. Association
  for Computational Linguistics.

\bibitem[{Glorot and Bengio(2010)}]{GlorotAISTATS2010}
Xavier Glorot and Yoshua Bengio. 2010.
\newblock Understanding the difficulty of training deep feedforward neural
  networks.
\newblock In \emph{JMLR W\&CP: Proceedings of the Thirteenth International
  Conference on Artificial Intelligence and Statistics (AISTATS 2010)},
  volume~9, pages 249--256.

\bibitem[{Junczys-Dowmunt(2018{\natexlab{a}})}]{junczys-dowmunt-2018-dual}
Marcin Junczys-Dowmunt. 2018{\natexlab{a}}.
\newblock \href {https://www.aclweb.org/anthology/W18-6478} {Dual conditional
  cross-entropy filtering of noisy parallel corpora}.
\newblock In \emph{Proceedings of the Third Conference on Machine Translation:
  Shared Task Papers}, pages 888--895, Belgium, Brussels. Association for
  Computational Linguistics.

\bibitem[{Junczys-Dowmunt(2018{\natexlab{b}})}]{junczys-dowmunt-2018-microsofts}
Marcin Junczys-Dowmunt. 2018{\natexlab{b}}.
\newblock \href {https://www.aclweb.org/anthology/W18-6415} {{M}icrosoft{'}s
  submission to the {WMT}2018 news translation task: How {I} learned to stop
  worrying and love the data}.
\newblock In \emph{Proceedings of the Third Conference on Machine Translation:
  Shared Task Papers}, pages 425--430, Belgium, Brussels. Association for
  Computational Linguistics.

\bibitem[{Junczys-Dowmunt and
  Grundkiewicz(2018)}]{junczys-dowmunt-grundkiewicz-2018-ms}
Marcin Junczys-Dowmunt and Roman Grundkiewicz. 2018.
\newblock \href {https://www.aclweb.org/anthology/W18-6467} {{MS}-{UE}din
  submission to the {WMT}2018 {APE} shared task: Dual-source transformer for
  automatic post-editing}.
\newblock In \emph{Proceedings of the Third Conference on Machine Translation:
  Shared Task Papers}, pages 822--826, Belgium, Brussels. Association for
  Computational Linguistics.

\bibitem[{Junczys-Dowmunt et~al.(2018)Junczys-Dowmunt, Grundkiewicz, Dwojak,
  Hoang, Heafield, Neckermann, Seide, Germann, Aji, Bogoychev, Martins, and
  Birch}]{junczys-dowmunt-etal-2018-marian-fast}
Marcin Junczys-Dowmunt, Roman Grundkiewicz, Tomasz Dwojak, Hieu Hoang, Kenneth
  Heafield, Tom Neckermann, Frank Seide, Ulrich Germann, Alham~Fikri Aji,
  Nikolay Bogoychev, Andr{\'e} F.~T. Martins, and Alexandra Birch. 2018.
\newblock \href {https://www.aclweb.org/anthology/P18-4020} {{M}arian: Fast
  neural machine translation in {C}++}.
\newblock In \emph{Proceedings of {ACL} 2018, System Demonstrations}, pages
  116--121, Melbourne, Australia. Association for Computational Linguistics.

\bibitem[{Kudo and Richardson(2018)}]{kudo-richardson-2018-sentencepiece}
Taku Kudo and John Richardson. 2018.
\newblock \href {https://www.aclweb.org/anthology/D18-2012} {{S}entence{P}iece:
  A simple and language independent subword tokenizer and detokenizer for
  neural text processing}.
\newblock In \emph{Proceedings of the 2018 Conference on Empirical Methods in
  Natural Language Processing: System Demonstrations}, pages 66--71, Brussels,
  Belgium. Association for Computational Linguistics.

\bibitem[{Post(2018)}]{post-2018-call}
Matt Post. 2018.
\newblock \href {https://www.aclweb.org/anthology/W18-6319} {A call for clarity
  in reporting {BLEU} scores}.
\newblock In \emph{Proceedings of the Third Conference on Machine Translation:
  Research Papers}, pages 186--191, Belgium, Brussels. Association for
  Computational Linguistics.

\bibitem[{Radford et~al.(2019)Radford, Wu, Child, Luan, Amodei, and
  Sutskever}]{radford2019language}
Alec Radford, Jeff Wu, Rewon Child, David Luan, Dario Amodei, and Ilya
  Sutskever. 2019.
\newblock Language models are unsupervised multitask learners.

\bibitem[{Tiedemann and Scherrer(2017)}]{tiedemann-scherrer-2017-neural}
J{\"o}rg Tiedemann and Yves Scherrer. 2017.
\newblock \href {https://doi.org/10.18653/v1/W17-4811} {Neural machine
  translation with extended context}.
\newblock In \emph{Proceedings of the Third Workshop on Discourse in Machine
  Translation}, pages 82--92, Copenhagen, Denmark. Association for
  Computational Linguistics.

\bibitem[{Vaswani et~al.(2017)Vaswani, Shazeer, Parmar, Uszkoreit, Jones,
  Gomez, Kaiser, and Polosukhin}]{NIPS2017_7181}
Ashish Vaswani, Noam Shazeer, Niki Parmar, Jakob Uszkoreit, Llion Jones,
  Aidan~N Gomez, \L~ukasz Kaiser, and Illia Polosukhin. 2017.
\newblock \href
  {http://papers.nips.cc/paper/7181-attention-is-all-you-need.pdf} {Attention
  is all you need}.
\newblock In I.~Guyon, U.~V. Luxburg, S.~Bengio, H.~Wallach, R.~Fergus,
  S.~Vishwanathan, and R.~Garnett, editors, \emph{Advances in Neural
  Information Processing Systems 30}, pages 5998--6008. Curran Associates, Inc.

\end{thebibliography}

\end{document}